\documentclass[11pt,a4paper,twocolumn]{article}
\usepackage[T1]{fontenc}
\usepackage[utf8]{inputenc}
\usepackage[english]{babel}
\usepackage{amsmath}
\usepackage{amsfonts}
\usepackage{appendix}
\usepackage{amssymb}
\usepackage{seqsplit}
\usepackage{array}
\newcolumntype{P}[1]{>{\centering\arraybackslash}p{#1}} 
\usepackage{url}
\usepackage{graphicx}
\usepackage{lscape}
\usepackage{supertabular}
\usepackage{hyperref}
\usepackage{verbatim}
\usepackage[left=2cm,right=2cm,top=3cm,bottom=3cm]{geometry}
\author{Matthew R. Karlsen and Sotiris Moschoyiannis}
\title{Customer Segmentation of Wireless Trajectory Data\\ \vspace{0.5cm}\small Technical Report, Department of Computer Science, University of Surrey, March 2018}
\begin{document}
\maketitle


\begin{abstract}
Wireless trajectory data consists of a number of (time, point) entries where each point is associated with a particular wireless device (WAP or BLE beacon) tied to a location identifier, such as a place name. A trajectory relates to a particular mobile device. Such data can be clustered `semantically' to identify similar trajectories, where similarity relates to non-geographic characteristics such as the type of location visited. Here we present a new approach to semantic trajectory clustering for such data. The approach is applicable to interpreting data that does not contain geographical coordinates, and thus contributes to the current literature on semantic trajectory clustering. The literature does not appear to provide such an approach, instead focusing on trajectory data where latitude and longitude data is available.

We apply the techniques developed above in the context of the Onward Journey Planner Application, with the motivation of providing on-line recommendations for onward journey options in a context-specific manner. The trajectories analysed indicate commute patterns on the London Underground. Points are only recorded for communication with WAP and BLE beacons within the rail network. This context presents additional challenge since the trajectories are `truncated', with no true origin and destination details. 

In the above context we find that there are a range of travel patterns in the data, without the existence of distinct clusters. Suggestions are made concerning how to approach the problem of provision of on-line recommendations with such a data set. Thoughts concerning the related problem of prediction of journey route and destination are also provided.
\end{abstract}

\section*{Acknowledgements}

The authors thank Innovate UK for partial funding of this work via the Accelerating Innovation in Rail 4 (AIR4) programme and CommuterHive for provision of the data set used. The authors also thank Stephan Wesemeyer for comments that have improved the quality of this report. Any errors or omissions are of course the authors' own.

\section{Introduction}

The ubiquity and computational power of modern smart phones makes phone-based travel assistance applications for rail users possible. The Onward Journey Planner Assistant (OJPA) is one such travel assistance application currently under construction. This project aims to go beyond current solutions to provide rail customers with suggestions for mode of travel when reaching the end of their rail journey or when faced with delays and cancellations to part of their rail journey. Such an application offers value to the customer by providing savings in terms of both time and money, as well as augmenting the customer's limited information on facilities in the surrounding area \cite{yoell-2017-onward}.

For effective travel assistance critical pieces of information must be known such as the current vehicle being used for the journey, other vehicles to be used within the current journey, and transport mode preferences. Travel assistance applications can themselves obtain much data. Whilst this data does not include such critical pieces of information per se, this data can be processed to infer some of the critical information required. The remainder of the information required must be collected from external services 
or via questions directed to the user.

Data collection on phone applications may be passive or active. Passive data collection includes the constant recording of GPS (latitude, longitude) data or data concerning when and where the phone enters the range of WAP devices or BLE beacons. It also involves instances where a user choice within the application (not a direct data gathering question) supplies us with information. Active data collection, in contrast, involves the application directly asking the user questions. Both passive and active data collection can improve the service that the application provides.

Critical information for a travel assistance application includes the destination for the current journey, the probable route they are currently taking, and the vehicles they will use on this route. Destination may be acquired, or at least confirmed, via active data collection. However, it is clearly preferred to gather such information passively to avoid unnecessarily disturbing the user. Either way, the route prediction must be gathered passively -- it is unrealistic to expect users to input the waypoints that are present on their journey. Once a predicted route is established, this information may be used in combination with rail APIs to work out which particular vehicles (particular services) a user will require on their journey. Further information such as the delay status of vehicles or services used or possible alternatives and their costs and speeds, may be obtained via external sources through the various APIs offered.
With the above information, when a delay or cancellation occurs on a currently used train or part of the passenger's onward journey, it will be possible to suggest one or more courses of action that the user can take to mitigate or partially mitigate the delay.

From the above we can see that route prediction is critical for improved travel assistance applications. However, it is also desirable to not only provide the user with recommendations but to customise these recommendations for the user. Customers may have different trade-off preferences concerning speed of journey, cost, noise and personal space, pollution impacts and 	disability access requirements. It may also be possible to include preferences that purely relate to mode of travel such as `car preference' within the `peterol head' consumer group identified in the `Traveller Needs and UK Capability Study' released by Transport Systems Catapult \cite{tsc-2015-traveller}. Whilst \emph{the passive locational data gathered is unable to directly inform on these preferences}, any selection of options within the application can be used to provide insight in to these preferences. For instance, if the user continually selects the fastest option, no matter the cost, we then know that the user has a strong preference concerning speed of journey. In contrast, if they consistently check for disabled access at stations, it is likely that the user has a physical disability they need to consider. These preferences, gathered through application use, can be used to tailor recommendations to the user.

One further question is whether clustering the passively collected locational data could reveal distinct clusters of users. If so, \emph{one could then examine whether there is a correspondence between these clusters and particular regions within the space of possible preferences}, through asking users in each locational cluster what their preferences are (possibly through a chat bot running on their phones). If and only if each cluster corresponds to a distinct region within the space of possible preferences, will it be possible to infer onward journey preferences from examination of the locational data. For this reason, a from of cluster analysis inspired by previous work on semantic trajectory clustering \cite{ying-2010-mining, parent-2013-semantic} has been applied to pre-processed locational data to explore whether distinct clusters (here, customer segments) exist within the locational data. If these distinct clusters exist, identification of preferences through processing of the gathered locational data is possible provided that information on user preferences is obtained for some representative subset of the user population is obtained.

\subsection{Planned Data Analysis}
\label{sect:daintro}

Customer segmentation is the process of dividing customers in to a number of different groups where each group has a different consumption pattern and different product preferences. Their particular non-consumption characteristics may also be very different. A good segmentation identifies groups such that within group similarity is very high (with respect to the features in question), whilst groups are as distinct as possible -- differences between each group are also very high.

Such segmentation is usually performed via data clustering or dimensionality reduction approaches. Clustering is the process of identifying distinct groups within an unlabelled data set. The aim of the data clustering is to minimise dissimilarity or maximise similarity within each cluster whilst maximising dissimilarity or minimising similarity between clusters. Clearly, whilst this process is simple in one or two dimensions, identifying groups within data becomes very difficult when there are four or more dimensions, thus necessitating a computational approach.

Some of the main approaches include partitional (e.g. K-Means), spectral, density-based (e.g. DBSCAN) and hierarchical clustering (e.g. via UPGMA) \cite{everitt-2001-cluster, jain-2010-data}. The usual input to these algorithms is an item-by-feature matrix and the output is a list of clusters with a list of items belonging to each cluster. Graphical representation of the clustering is usually not provided and must instead be produced via other means.

Some algorithms produce alternative output. For instance, hierarchical clustering produces a dendrogram (hierarchy diagram) of the data. To produce clusters from this, a cut-point must then be chosen such that the hierarchical output is converted to flat clusters.

Here we wish to perform clustering on trajectory data in order to identify distinct groups of travellers based upon their travel patterns.\footnote{We emphasise that our approach is inspired by current approaches to \emph{semantic} trajectory clustering, rather than `regular' geographic trajectory clustering. Regular trajectory clustering aims to find geographically similar routes and is thus used for applications such as prediction \cite{chen-2011-personal}. In contrast, semantic trajectory clustering aims to find trajectories with similar patterns concerning the types of location visited \cite{ying-2010-mining}, without discriminating based on geographic location.} Such data may be acquired in a number of ways, the most common being GPS. The data source in this paper involves smartphone-holding customers with an application installed on their phone that records their entry and exit interactions with wireless access points (WAP) and/or Bluetooth Low Energy (BLE) beacons. The data acquired from these interactions records the trajectory of the customer through the areas where the WAPs and BLE beacons exist. The data only includes WAPs and BLE beacons in railway stations and not elsewhere. Each WAP or BLE beacon is associated with a region (e.g. station turnstiles). In turn, each region is associated with a location (such as a particular station). Associated data on the mobile phone specification is also collected.

Because wireless trajectory data is not in an item-by-feature format, conventional clustering methods cannot be applied to such data without either modifying the algorithms used or preprocessing the data. (The same is true of time series data.) Trajectory clustering and data mining methods exist \cite{lee-2007-trajectory,zheng-2015-trajectory}. However, regular trajectory clustering methods usually cluster `geographically' (users with similar routes in geographic space are clustered together). This is useful for prediction. Here, however, we are instead seeking a customer segmentation such that users with similar route patterns are clustered together irrespective of their geographic position. For instance, a commuter who travels 50 minutes in to London by train from the same stop each day, to the same stop each day, should be considered very similar to another commuter who commutes to Manchester with the same pattern (and different start and end stops). Such a goal is strongly related to the field of `semantic trajectory clustering' \cite{parent-2013-semantic}, which relates each stop with some kind of meaning (e.g. home, work, medical facility, supermarket, etc) and then clusters the semantically-annotated trajectories. Features may also be derived from the trajectory data and clustering may be performed on the derived features.

Here we use a hybrid approach, first obtaining features through pre-processing, and then clustering using both these features and a semantically-annotated journey pattern using DBSCAN \cite{ester-1996-density}. This hybrid approach is necessary because of the type of data that we have. The data contains no private locations (thus excluding trajectory starts and ends) and contains no locations other than stations (and thus full semantic annotation is not possible). 

The data is also incomplete for a number of behavioural and technical reasons that are difficult to mitigate:

\begin{itemize}
\item users may not always carry their phone
\item the phone may not always be on
\item the phone may not be running the relevant application
\item the application may have been terminated by the host operating system
\item the phone operating system may not be permitting the phone to connect to WAPs or BLE beacons
\item the phone operating system may not permit the phone to transmit the collected data
\end{itemize} 
Processes must therefore be developed that can cope with data that may be incomplete. There is also no `training data set' with which to evaluate the accuracy of the preprocessing steps (we have to rely on human inspection). This makes evaluating adjustments to the preprocessing steps difficult. Finally, due to wireless `quirks', sometimes entry in to a wireless area is recorded whilst the exit is not recorded. When the user re-enters the wireless area later in the day, and then exits, the exit may be recorded along side the entrance time from earlier in the day, thereby making it seem that the user has remained in the station for a very long interval of time. The data pre-processing steps must be designed to take this issue in to account.

\subsection{Software Used}

The software libraries in Table \ref{tab:sl} were used to process the data and produce the figures and table information used in this report. The report itself was produced using the TeX Live \LaTeX distribution.

\begin{table*}
\small
\begin{tabular}{|c|c|c|c|c|c|}
\hline 
\textbf{Library Group ID} & \textbf{Library Artifact ID} & \textbf{Version} & \textbf{License} & \textbf{Purpose} \\ 
\hline 
\hline
org.postgresql & postgresql & 42.1.4 & BSD 2-Clause & Database interaction \\ 
\hline 
org.apache.logging.log4j & log4j-api & 2.10.0 & Apache 2.0 & Logging \\ 
\hline 
joda-time & joda-time & 2.9.9 & Apache 2.0 & Time calculations \\ 
\hline 
org.apache.logging.log4j & log4j-core & 2.10.0 & Apache 2.0 & Logging \\ 
\hline 
org.apache.commons & commons-csv & 1.5 & Apache 2.0 & Spreadsheed writing \\ 
\hline 
com.github.haifengl & smile-core & 1.5.0 & Apache 2.0 & Clustering \\ 
\hline 
de.erichseifert.gral & gral-core & 0.11 & LGPL & Chart production \\ 
\hline 
junit & junit & 4.12 & EPL 1.0 & Testing \\ 
\hline 
org.mockito & mockito-core & 2.13.0 & MIT & Testing \\ 
\hline 
com.google.code.gson & gson & 2.8.2 & Apache 2.0 & Data file writing/retrieval \\ 
\hline 
commons-io & commons-io & 2.6 & Apache 2.0 & Efficient file I/O \\ 
\hline 
org.apache.commons & commons-text & 1.2 & Apache 2.0 & Additional distance measures \\ 
\hline 
org.apache.commons & commons-math3 & 3.6.1 & Apache 2.0 & Principal components analysis \\ 
\hline 
org.apache.httpcomponents & httpclient & 4.5.4 & Apache 2.0 & HTTP requesting \\
\hline
org.apache.httpcomponents & fluent-hc & 4.5.4 & Apache 2.0 & HTTP requesting \\
\hline
\end{tabular}
\caption{The software libraries used to produce this report.}
\label{tab:sl}
\end{table*}

\subsection{Report Layout}

The remainder of the report proceeds as follows. Section \ref{sect:pp} explains the pre-processing. Scatter-plots (Section \ref{sect:scatter}) and principal component analysis (Section \ref{sect:drdv}) are then used to gain low dimensionality visualisations of the data. The clusering process is then described in Section \ref{sect:dc}. In Section \ref{sect:results} the results of the DBSCAN clustering are then overlaid on the PCA output via colouring each data point according to the cluster it is a member of. Future work is presented in Section \ref{sect:fwork}.

\section{Data Preprocessing}
\label{sect:pp}

\subsection{Introduction}
\label{sect:ppint}

Upon inspection of the trajectory data it becomes apparent that some useful customer `features' can be extracted via a `pre-processing' stage. Other features (such as the details of user mobile phones) may be available without further processing. There are also features that may be obtained through combining the data set with other sources of information such as a model of the rail network. The various potential features include the following:

\begin{itemize}
\item total time spent travelling
\item average journey duration
\item total distance of travel (direct or the distance along the route actually travelled)
\item distance of each journey
\item the times of day person is travelling
\item the journey pattern (work to home, home to work, ...)
\item the phone type of the user
\item number of locations visited per journey
\item journey frequency
\item number of journeys per `offline rest location'\footnote{See Section \ref{sect:orl} for an explanation of offline rest locations.}
\item distance travelled per unit of time
\end{itemize}

\subsection{Preliminary Steps}

The first step applied in processing the wireless data is to split each `entry and exit time' pair in to individual points. This is necessary because wireless data overlaps. For instance, a mobile may come in range of a WAP and then come in range of a BLE beacon, then leave the BLE beacon and then leave the WAP. If overlapping data is simply filtered out then a large number of correct data points may be removed, for this reason we argue that splitting the entry and exit times to individual points (entry points and exit points) is necessary before we proceed. When considering the WAP and BLE beacon example above, this means that the two initial points (each containing both an entry and exit time) are split in to four separate points (though they are still related by the original point ID within the pre-processing program).

Some preliminary filtering may be performed at this time. If the exit time at point 1 is later than the entry time at point 2 then point 2 is deemed problematic and is discarded \emph{if} point 2 is at a different location than point 1 (otherwise the points are just split). This filtering is applied to remove any points that would require that the customer be in two locations at once.

\subsection{Journey Extraction}
\label{sect:journeyext}

At this stage it is now possible to identify possible journeys. The first chronological point is added to the first journey. We iterate through each chronological pair, starting with the first and second points. The time difference between the points is considered. If they are more than 80 minutes apart then they are considered to belong to separate journeys and a new journey is started with the second point being considered. The iteration through each chronological pair proceeds in this way until the entire trajectory has been broken up in to journeys.

The 80 minute threshold is selected because one can be reasonably certain that after 80 minutes a journey on the London Underground will have finished. Clearly if this threshold is set too low then additional erroneous journeys may be created. On the other hand, if set too high, journeys may be erroneously merged in to one another. Most journeys on the underground are relatively short, and thus the 80 minute threshold may seem high. However, the longest journey on the Underground, from Epping to West Ruislip on the Central line \cite{tfl-2018-facts}, takes 1 hour and 27 minutes \cite{tfl-2018-times}. 80 minutes ensures the vast majority of journeys will not be split incorrectly. The limitations of this approach are considered further in Section \ref{sect:fwork}.

\subsection{Offline Rest Locations}
\label{sect:orl}

Once possible journeys have been identified, offline rest locations (ORLs) can be identified. Offline rest locations are similar to the concept of a ``stop'' \cite{xiang-2016-extracting} or ``stay point'' \cite{zheng-2009-mining} within the literature on trajectory analysis. An offline rest location is simply an exit from and entry to a particular location\footnote{ORLs are calculated based on location, not a particular WAP, BLE beacon, or region. It is possible that a user exits from one platform and re-enters at another platform where each platform has a different associated WAP and region.} where the time difference between the two is greater than a particular threshold value. The threshold value used here is 30 minutes. This 30 minute threshold is set to a precautionary level to avoid creating false ORLs when users enter and exit wireless `black spots' on the platform or in surrounding station facilities. It is unlikely that a user will be in a black spot for more than a few minutes and if the user exits and then enters a location with a time gap of greater than 30 minutes it is relatively safe to assume they have actually left the station.

At this stage additional filtering is performed. Consumer \emph{instances with trajectories that contain fewer than ten points are discarded} because they are deemed to not be sufficiently detailed to accurately give an idea of that user's overall usage pattern. Instances with a trajectory duration less than 24 hours are also discarded for similar reasons. Finally, those points with timestamps earlier than 2000-01-01 00:00 are also removed. Of the 348,304 items that have one or more data points, 98,601 pass these filters.\footnote{One additional filter was later added to remove points with erroneous timestamps. Some of the trajectories began in 1970 and thus were clearly invalid. A filter was introduced to remove all trajectories that began before the start of the year 2000.}

\subsection{Labelling Offline Rest Locations}
\label{sect:trl}

It is possible that the offline rest locations identified above have special meaning to the customer in question. For instance, they could be a \emph{home station} (the station closest to the person's home) or they could be a \emph{work station} (the station closest to the person's place of work). Here we apply a simple approach that attempts to identify the possible home and work stations.

Each offline rest location (and therefore possible home or work station) is assessed according to entry times, exit times, frequency of use and duration. A `home score' and `work score' is associated with each offline rest location, determined by those factors just mentioned. Work exit times are assessed first. The most frequent depart time is obtained from each offline rest location and if the most frequent time lies between 17:00 and 21:00 then the work score for that offline rest location is increased by 3. A similar process is applied for arrival times, except the hour range that triggers the work score increment is 05:00 to 10:00. Home entry and exit times are again similar but use the time ranges 17:00 to 21:00 and 05:00 to 10:00 respectively (and it is the home score rather than the work score which is increased).

The score modification for frequency of use and duration is performed as follows. The two offline rest locations with the longest rest durations have their home and work scores incremented by 2. The two most frequently used offline rest locations also have their home and work scores incremented by 2.

The values 2 and 3 (above) were chosen such that an ORL must be both the most frequently used and be used for the longest duration before such an ORL is seen as more important than one where the commute times are feasible (as determined by the mechanism above).

Once scoring is complete, the home offline rest location is selected as that with the highest home score. The work offline rest location is then selected as that with the highest work score, excluding the ORL selected as the home location.

\subsection{Journey Pattern}
\label{sect:jpatt}

For each consumer, for each journey, we create a journey pattern, which will be used in the clustering process later. A journey pattern consists of a number of tokens, denoting work station, home station, unknown station and offline rest location. So, annotating locations in the journey pattern as (H)home, (W)ork, (U)nknown, or (O)ffline rest location, a particular journey might look like \texttt{H,U,W} or \texttt{H,U,O}. A count is created for each journey pattern such that we have a map, relating each unique journey pattern (for a given consumer) with the number of times that pattern has been used by the consumer in question.

\subsection{Manual Examination of Pre-Processed Output}

The data pre-processing steps described above work well in many cases. In some other cases they are unable to extract useful information from the data. In Section 1 of the supplement to this report \cite[Section 1]{karlsen-2018-supplement}, five commuter patterns (\emph{referred to as Commuter 1 to Commuter 5}) that were successfully extracted from the data are presented. In Section 2 of the supplement \cite[Section 2]{karlsen-2018-supplement}, five problematic patterns of different types are listed (referred to as Problematic Pattern 1 to Problematic Pattern 5). In this section we will first discuss the successful patterns. We will then examine the problematic patterns and explain why each of them is problematic.

\subsubsection{Commuter Trajectories 1 to 5}

We first show the journey string of Commuter 1 and the abbreviated form of the output from the pre-processing. The journey string is as follows: \begin{quotation}\noindent\seqsplit{H|H|HW|W|HW|U|H|W|W|UWUH|H|UU|HU|HW|HW|WH|HUW|HW|UW|HW|W|W|HW|WH|HW|UW|W|WH|HW|HW|UH|HW|WH|HW|UH|HW|H|WH|HW|HW|H|U|H|W|H|HW|UW|HW|WUH|UUW|WU|H|UWUUUH|H|H|H|H|H|H|HW|UW|H|W|H|HW|WH|HUW|H|H|H|H|H|W|W|H|H|H|H|H|H|HW|U|H|H|H|U|H|H|H|HW|H|H|H|H|W|H|U|HW|W|H|H|W|H|U}\end{quotation} where H represents use of the `home station', W represents use of the `work station' and U represents an unlabelled location. The vertical bars separate the distinct journeys. It is clear that some of the journeys consist of a single location. Single location journeys are due to the behavioural and technical reasons listed in Section \ref{sect:daintro}. A condensed journey list from the preprocessing is shown in Table \ref{tab:mi1}. The ORLs for commuter 1 are Canning Town, with the following durations (in hours):\begin{quotation}\noindent335.98; 8.4; 24.08; 8.44; 8.6; 8.44; 8.45; 8.18; 8.48; 8.32; 15.6; 8.69\end{quotation} and London Bridge, with durations:\begin{quotation}\noindent23.67; 24.05; 71.95; 23.98; 23.97; 23.86; 48.13; 71.99; 8.93; 62.95; 47.98; 33.21; 119.72; 23.96; 48; 23.85; 32.77; 32.87; 815.99; 23.94; 23.93; 144.05; 95.78.\end{quotation}The pre-processing has labelled the most likely home station as London Bridge Station and the most likely work station as Canning Town Station. We see from visual inspection of Table \ref{tab:mi1} (and the full pre-processed output) that there is a high number of journeys from London Bridge to Canning Town in the morning and several journeys from Canning Town to London Bridge in the afternoon. Pre-processing therefore appears to have labelled work and home correctly in this instance.

\begin{table}
\begin{center}
\begin{tabular}{p{5cm} P{2cm}}
\small
Journey & Count \\
\hline
\hline
London Bridge Station & AM (42), PM (0) \\
\hline
London Bridge Station \(\rightarrow\) Canning Town Station & AM (22), PM (0) \\
\hline
Canning Town Station & AM (0), PM (13) \\
\hline
Canning Town Station \(\rightarrow\) London Bridge Station & AM (0), PM (6) \\
\hline
West Ham Station & AM (0), PM (4) \\
\hline
West Ham Station \(\rightarrow\) Canning Town Station & AM (0), PM (4) \\
\hline
West Ham Station \(\rightarrow\) Canning Town Station \(\rightarrow\) North Greenwich Station \(\rightarrow\) London Bridge Station & AM (0), PM (1) \\
\hline
North Greenwich Station & AM (0), PM (1) \\
\hline
Canning Town Station \(\rightarrow\) Canary Wharf \(\rightarrow\) London Bridge Station & AM (0), PM (1) \\
\hline
London Bridge Station \(\rightarrow\) North Greenwich Station & AM (1), PM (0) \\
\hline
London Bridge Station \(\rightarrow\) Bermonsey Station \(\rightarrow\) Canning Town Station & AM (1), PM (0) \\
\hline
Bermonsey Station \(\rightarrow\) London Bridge Station & AM (0), PM (1) \\
\hline
London Bridge Station \(\rightarrow\) North Greenwich Station \(\rightarrow\) Canning Town Station & AM (1), PM (0) \\
\hline
Waterloo \(\rightarrow\) North Greenwich Station \(\rightarrow\) Canning Town Station & AM (1), PM (0) \\
\hline
Canning Town Station \(\rightarrow\) Bermonsey Station & AM (0), PM (1) \\
\hline
West Ham Station \(\rightarrow\) Canning Town Station \(\rightarrow\) North Greenwich Station \(\rightarrow\) Canary Wharf \(\rightarrow\) Bermonsey Station \(\rightarrow\) London Bridge Station & AM (0), PM (1) \\
\hline
...
\end{tabular} 
\end{center}
\caption{The condensed journey list of the first commuter. The supplement \cite[Section 1.1]{karlsen-2018-supplement} displays the full pre-processing output for this commuter.}
\label{tab:mi1}
\end{table}

Commuter 2 presents a more difficult instance because there are few journeys of two or more locations to work with (the majority of `journeys' are a collection of points at one single location or another). The journey string is as follows:\begin{quotation}\noindent\seqsplit{W|O|OH|HW|W|OH|HW|W|W|WO|HOW|H|W|H|OH|HW|WOHUH|W|WUOH|H|W|HW|WHU|H|OH|OW|WO|U|U|HW|W|H|OH|O|H|HW|WH|OW|H}\end{quotation}The offline rest locations are at Tower Hill (duration 7.72), Bank \& Monument Station with durations:\begin{quotation}16.19; 15.24; 72.23; 14.88\end{quotation} and at Aldgate East Station (durations 25.79; 3.36; 47.42). Despite the difficulty of the example, the pre-processing has identified a home station and work station. Inspecting the condensed pre-processed output shown in Table \ref{tab:mi2} supports the identification of the home station as Bank \& Monument Station and the work station as Aldgate East Station.

\begin{table}
\begin{center}
\begin{tabular}{p{5cm} P{2cm}}
\small
Journey & Count \\
\hline
\hline
Aldgate East Station & AM (8), PM (0) \\
\hline
Bank and Monument Station & AM (0), PM (7) \\
\hline
Bank and Monument Station \(\rightarrow\) Aldgate East Station & AM (6), PM (0) \\
\hline
Tower Hill \(\rightarrow\) Bank and Monument Station & AM (0), PM (5) \\
\hline
Tower Hill \(\rightarrow\) Aldgate East Station & AM (2), PM (0) \\
\hline
Aldgate East Station \(\rightarrow\) Tower Hill & AM (0), PM (2) \\
\hline
Tower Hill & AM (2), PM (0) \\
\hline
Aldgate East Station \(\rightarrow\) Bank and Monument Station & AM (0), PM (1) \\
\hline
Oxford Circus & AM (1), PM (0) \\
\hline
Aldgate East Station \(\rightarrow\) Bank and Monument Station \(\rightarrow\) London Bridge Station & AM (0), PM (1) \\
\hline
Aldgate East Station \(\rightarrow\) Aldgate Station \(\rightarrow\) Tower Hill \(\rightarrow\) Bank and Monument Station & AM (0), PM (1) \\
\hline
Bank and Monument Station \(\rightarrow\) Tower Hill \(\rightarrow\) Aldgate East Station & AM (1), PM (0) \\
\hline
Victoria & AM (0), PM (1) \\
\hline
Aldgate East Station \(\rightarrow\) Tower Hill \(\rightarrow\) Bank and Monument Station \(\rightarrow\) London Bridge Station \(\rightarrow\) Bank and Monument Station & AM (1), PM (0) \\
\end{tabular} 
\end{center}
\caption{The condensed journey list of the second commuter. The supplement \cite[Section 1.2]{karlsen-2018-supplement} displays the full pre-processing output for this commuter.}
\label{tab:mi2}
\end{table}


Commuter 3 commutes from Highbury \& Islington Station to Victoria. The journey string is:\begin{quotation}\noindent\seqsplit{H|H|HW|HU|H|U|HU|O|W|U|H|H|HU|HUW|HUW|H|O|U|W|WU|UUU|U|HW|HW|H|WH|UO|U|WH|O|O|WH|U|UO|O|UO|O|HW|WH|HW|WH|O|WH|WH|HW|UH|HW|W|W|H|O|W|HWUUO|W|H|W|H|HW|WH|W|O|HW|OU|W}\end{quotation} The ORLs are Highbury \& Islington Station with durations:\begin{quotation}\noindent672.6; 34.82; 10.84; 72.46; 10.53; 11.22\end{quotation} Victoria with durations:\begin{quotation}\noindent23.8; 48.08; 12.2; 11.5; 12.35\end{quotation} and Holloway Road Station (durations 2.16; 9.73; 18.45). The most likely home station was identified as Highbury \& Islington Station, whilst the most likely work station was identified as Victoria. Inspection of the condensed pre-processed data in Table \ref{tab:mi3} supports this labelling. It should be noted that the phone in question often also connects at Kings Cross St. Pancras Underground station, part-way along the journey. Connections at Warren Street are also possible. This commuter also has an additional ORL at Holloway Road Station.

\begin{table}
\begin{center}
\begin{tabular}{p{5cm} P{2cm}}
\small
Journey & Count \\
\hline
\hline
Highbury and Islington Station & AM (0), PM (10) \\
\hline
Highbury and Islington Station \(\rightarrow\) Victoria & AM (9), PM (0) \\
\hline
Victoria & AM (0), PM (9) \\
\hline
Holloway Road Station & AM (0), PM (9) \\
\hline
Victoria \(\rightarrow\) Highbury and Islington Station & AM (8), PM (0) \\
\hline
Knightsbridge & AM (3), PM (0) \\
\hline
Highbury and Islington Station \(\rightarrow\) Kings Cross and St Pancras \(\rightarrow\) Victoria & AM (2), PM (0) \\
\hline
Warren Street Station & AM (2), PM (0) \\
\hline
Highbury and Islington Station \(\rightarrow\) Warren Street Station & AM (2), PM (0) \\
\hline
Kings Cross and St Pancras \(\rightarrow\) Notting Hill Gate Station \(\rightarrow\) Shepherd's Bush & AM (0), PM (1) \\
\hline
Holloway Road Station \(\rightarrow\) Knightsbridge & AM (0), PM (1) \\
\hline
Knightsbridge \(\rightarrow\) Holloway Road Station & AM (1), PM (0) \\
\hline
Highbury and Islington Station \(\rightarrow\) Kings Cross and St Pancras & AM (0), PM (1) \\
\hline
Highbury and Islington Station \(\rightarrow\) Victoria \(\rightarrow\) Brixton \(\rightarrow\) Kings Cross and St Pancras \(\rightarrow\) Holloway Road Station & AM (1), PM (0) \\
\hline
Pimlico Station \(\rightarrow\) Highbury and Islington Station & AM (0), PM (1) \\
\hline
Holborn Station & AM (0), PM (1) \\
\hline
Victoria \(\rightarrow\) Warren Street Station & AM (1), PM (0) \\
\hline
...
\end{tabular} 
\end{center}
\caption{The condensed journey list of the third commuter. The supplement \cite[Section 1.3]{karlsen-2018-supplement} displays the full pre-processing output for this commuter.}
\label{tab:mi3}
\end{table}

Commuter 4 has regular patterns of WUH or HUW. The journey string is:\begin{quotation}\noindent\seqsplit{HW|WUH|WU|WU|HUW|W|HUW|UUU|H|UU|UUUUH|HUW|HUUUU|UUUUH|HU|HUW|WUH|HUUUUU|UH}\end{quotation} The ORLs are Clapham Common (durations 16.47; 9.88; 15.2) and Victoria (duration 8.49). This commuter joins the underground network at Clapham Common (labelled as home) and travels on the Northern line to Stockwell (the intermediate stop in their journey pattern). From Stockwell they take the Victoria line to Victoria (labelled as work).  This labelling is tentatively supported by Table \ref{tab:mi4}.

\begin{table}
\begin{center}
\begin{tabular}{p{5cm} P{2cm}}
\small
Journey & Count \\
\hline
\hline
Clapham Common \(\rightarrow\) Pimlico Station \(\rightarrow\) Victoria & AM (4), PM (0) \\
\hline
Victoria \(\rightarrow\) Stockwell \(\rightarrow\) Clapham Common & AM (0), PM (2) \\
\hline
Victoria \(\rightarrow\) Stockwell & AM (0), PM (2) \\
\hline
Bank and Monument Station \(\rightarrow\) London Bridge Station \(\rightarrow\) Elephant and Castle Station \(\rightarrow\) Kennington \(\rightarrow\) Clapham Common & AM (0), PM (2) \\
\hline
Clapham Common \(\rightarrow\) Stockwell \(\rightarrow\) Kennington \(\rightarrow\) Waterloo \(\rightarrow\) Embankment \(\rightarrow\) Westminister Station & AM (1), PM (0) \\
\hline
Clapham Common \(\rightarrow\) Victoria & AM (0), PM (1) \\
\hline
Clapham Common & AM (0), PM (1) \\
\hline
St James' Park Station \(\rightarrow\) Clapham Common & AM (0), PM (1) \\
\hline
Russel Square Station \(\rightarrow\) Bank and Monument Station \(\rightarrow\) Stockwell & AM (0), PM (1) \\
\hline
Chancery Lane Station \(\rightarrow\) Holborn Station & AM (0), PM (1) \\
\hline
Victoria & AM (0), PM (1) \\
\hline
Clapham Common \(\rightarrow\) Kennington \(\rightarrow\) Waterloo \(\rightarrow\) Camden Town Station \(\rightarrow\) Kentish Town & AM (0), PM (1) \\
\hline
Clapham Common \(\rightarrow\) Clapham North & AM (1), PM (0) \\
\end{tabular} 
\end{center}
\caption{The condensed journey list of the fourth commuter. The supplement \cite[Section 1.4]{karlsen-2018-supplement} displays the full pre-processing output for the fourth commuter.}
\label{tab:mi4}
\end{table}

Commuter 5 is a more complex example, where the phone often manages to connect to WAP points and BLE beacons on part of the route, but not the whole route. The journey string is:\begin{quotation}\noindent\seqsplit{OUUW|HOUUW|W|UUW|WUUO|HUUW|WUHUH|UUW|W|HUW|W|WUO|OUUW|WUUH|HOUUW|WUUOH|OUUW|HUUUW|WUUOH|HOUUUW|WUH|W|WUUOH|UW|HUUW|WUOH|UUUW|WUUUOH|HUUUW|WUUUOH|HOUUW|H|HOUUW|WUUUH|HUUW|WUUUH|HOUUW|WUUUOH|HOW|WUUUOH|HUUW|HOW|W|WUUUOH|HOUW|WUUOH|HUUW|WUUOH|HOUW|UUH}\end{quotation} The ORLs are Piccadilly Station with durations:\begin{quotation}\noindent8.61; 9.48; 24.12; 8.82; 8.54; 9.55; 8.42; 8.56; 7.72; 8.36; 9.3; 6.28; 10.11; 9.04; 25.21; 8.66; 8.98; 9.96\end{quotation} Russel Square Station (duration 38.84) and Kings Cross \& St Pancras with durations:\begin{quotation}\noindent14.81; 14.99; 14.61; 14.54; 61.97; 13.75; 14.15\end{quotation} Visual inspection of the journeys (see Table \ref{tab:mi5}) suggests a most likely home station of Kings Cross St. Pancras and a most likely work station of Piccadilly Station -- the customer is commuting on the Piccadilly line.

Examination of the pre-processed output indicates that the home and work locations have been successfully identified, despite misleading shorter journey patterns such as Holborn, Leicester Square, Piccadilly (missing out the earlier stops and the Covent Garden stop). Commuter 5 is an example of why it is important to consider offline rest locations when labelling locations as work and home. \emph{Without the intermediate step of ORL-creation it is quite possible that the work and home labels could be attached to one or more intermediate stops}.

\begin{table}
\begin{center}
\begin{tabular}{p{5cm} P{2cm}}
\small
Journey & Count \\
\hline
\hline
Piccadilly Station \(\rightarrow\) Leicester Square Station \(\rightarrow\) Covent Garden Station \(\rightarrow\) Holborn Station \(\rightarrow\) Russel Square Station \(\rightarrow\) Kings Cross and St Pancras & AM (0), PM (5) \\
\hline
Piccadilly Station & AM (5), PM (0) \\
\hline
Kings Cross and St Pancras \(\rightarrow\) Russel Square Station \(\rightarrow\) Covent Garden Station \(\rightarrow\) Leicester Square Station \(\rightarrow\) Piccadilly Station & AM (3), PM (0) \\
\hline
Piccadilly Station \(\rightarrow\) Leicester Square Station \(\rightarrow\) Holborn Station \(\rightarrow\) Russel Square Station \(\rightarrow\) Kings Cross and St Pancras & AM (0), PM (3) \\
\hline
Kings Cross and St Pancras \(\rightarrow\) Covent Garden Station \(\rightarrow\) Leicester Square Station \(\rightarrow\) Piccadilly Station & AM (3), PM (0) \\
\hline
Russel Square Station \(\rightarrow\) Covent Garden Station \(\rightarrow\) Leicester Square Station \(\rightarrow\) Piccadilly Station & AM (3), PM (0) \\
\hline
Piccadilly Station \(\rightarrow\) Leicester Square Station \(\rightarrow\) Covent Garden Station \(\rightarrow\) Russel Square Station \(\rightarrow\) Kings Cross and St Pancras & AM (0), PM (2) \\
\hline
Piccadilly Station \(\rightarrow\) Leicester Square Station \(\rightarrow\) Covent Garden Station \(\rightarrow\) Holborn Station \(\rightarrow\) Kings Cross and St Pancras & AM (0), PM (2) \\
\hline
Kings Cross and St Pancras \(\rightarrow\) Russel Square Station \(\rightarrow\) Piccadilly Station & AM (2), PM (0) \\
\hline
...
\end{tabular} 
\end{center}
\caption{The condensed journey list of the fifth commuter. The supplement \cite[Section 1.5]{karlsen-2018-supplement} displays the full pre-processing output for the fifth commuter.}
\label{tab:mi5}
\end{table}

\subsubsection{Problematic Trajectories 1 to 5}

We now discuss the problematic trajectories from Supplement Section 2 \cite[Section 2]{karlsen-2018-supplement}. Problematic Trajectory 1 is an example of an ambiguous trajectory. The journey string is as follows: \begin{quotation}\noindent\seqsplit{OUUUUH|OUUUO|UUH|HUUO|OUUUUUUUH|H|HUU|O|HUUW|W|HUU|UH|W|UUUW|HUUUUUU|U|HUUW|U|UUH|HUUWUH|H|WUUH|HUW|WU|HUUW|HUUUW|WUU|U|UUUW|W|HUU|HUUU|UUW|WUUUH|HUUW|U|HUU|UUH|HUUUW|WUU|HUUUW|UUH|HUUW|UUH|HU|HUUUUUUUH|O|O|O|HUUOUU|O|HUWUU|UW|WUUH|HUU|U|HUW|WU|HUUUUUO|OUUU|HUU|HU|WUUUH|HU|WUUH|UUUUUUO|O|HUUUUUUO|OUUUUH|HUUUW|UUUH|HUU|HUUUU|WU|UW|WUH|UUH|H|WUUHW|WUUUH|U|WH|W|WUH}\end{quotation} The condensed journey list is shown in Table \ref{tab:pt1}. The ORLs for Problematic Trajectory 1 are Stepney Green, with the following durations (in hours):\begin{quotation}\noindent41.55; 235.46; 18.19; 15.92; 20.35; 13.25; 15.09; 72.55; 1.38; 1.49; 10.79; 1.47\end{quotation} Stockwell, with durations:\begin{quotation}\noindent5.86; 4.97; 1.75; 4.19\end{quotation} Plaistow Station with durations:\begin{quotation}\noindent1.83; 74.87; 4.05; 1.68; 3.9; 3.79\end{quotation} and Seven Sisters with duration 68.24. The pre-processing has labelled the most likely home station as Stepney Green Station and the most likely work station as Plaistow Station. On balance, this would appear to be correct. However, there is substantial uncertainty in this prediction, since there are few multi-location journeys that are used more than a few times (see Table \ref{tab:pt1}). We must therefore be aware of the possibility that this labelling is incorrect.

\begin{table}
\begin{center}
\begin{tabular}{p{5cm} P{2cm}}
\small
Journey & Count \\
\hline
\hline
Seven Sisters Station & AM (0), PM (5) \\
\hline
Plaistow Station & AM (0), PM (4) \\
\hline
Stepney Green \(\rightarrow\) Bow Road Station \(\rightarrow\) Bromley-by-Bow Station \(\rightarrow\) West Ham Station \(\rightarrow\) Plaistow Station & AM (3), PM (0) \\
\hline
Stepney Green & AM (0), PM (3) \\
\hline
Stepney Green \(\rightarrow\) Mile End Stn \(\rightarrow\) Stratford Station & AM (0), PM (3) \\
\hline
West Ham Station \(\rightarrow\) Mile End Stn \(\rightarrow\) Stepney Green & AM (0), PM (2) \\
\hline
Stratford Station \(\rightarrow\) Mile End Stn \(\rightarrow\) Stepney Green & AM (0), PM (2) \\
\hline
Plaistow Station \(\rightarrow\) Mile End Stn \(\rightarrow\) Stepney Green & AM (0), PM (2) \\
\hline
Stepney Green \(\rightarrow\) Mile End Stn \(\rightarrow\) Plaistow Station & AM (1), PM (0) \\
\hline
Stepney Green \(\rightarrow\) Stratford Station & AM (0), PM (1) \\
\hline
Bromley-by-Bow Station \(\rightarrow\) Plaistow Station & AM (1), PM (0) \\
\hline
Stepney Green \(\rightarrow\) Mile End Stn \(\rightarrow\) Bromley-by-Bow Station \(\rightarrow\) Plaistow Station & AM (0), PM (1) \\
\hline
Stepney Green \(\rightarrow\) Bow Road Station \(\rightarrow\) Plaistow Station \(\rightarrow\) Bow Road Station \(\rightarrow\) Mile End Stn & AM (1), PM (0) \\
\hline
West Ham Station \(\rightarrow\) Bow Road Station \(\rightarrow\) Mile End Stn \(\rightarrow\) Stepney Green & AM (1), PM (0) \\
\hline
Stepney Green \(\rightarrow\) Mile End Stn \(\rightarrow\) West Ham Station \(\rightarrow\) Plaistow Station & AM (1), PM (0) \\
\hline
Stepney Green \(\rightarrow\) Tower Hill \(\rightarrow\) Bank and Monument Station \(\rightarrow\) Stockwell \(\rightarrow\) Liverpool Street \(\rightarrow\) Leyton Station & AM (0), PM (1) \\
\hline
...
\end{tabular} 
\end{center}
\caption{The condensed journey list of Problematic Trajectory 1. The supplement \cite[Section 2.1]{karlsen-2018-supplement} displays the full pre-processing output for this trajectory.}
\label{tab:pt1}
\end{table}

Problematic Trajectory 2 shows an issue of offline rest location mis-labelling. The journey string is as follows: \begin{quotation}\noindent\seqsplit{U|OW|UHUW|U|UHW|WUU|WU|W|U|W|U|UH|UW|U|WU|U|U|UUUU|H|OU|UU|U|U|W|W|W|WUO|U|WUU|HW|U|W|UH|UW|UO|H|W|U|U|UUH|HUUUU|WU|UW|W|W|H|HW|U|U|W|WU|HU|U|U|H|WH|UUUU|UH|U|O|OUU|UU|H|U|UW|UUU|W|O|U|UW|W|W|O|U|UU|UU|HW|WH|UU|H|W|WHO|U|W|U|UOHU|U|O}\end{quotation} The abbreviated form of the output from the preprocessing is shown in Table \ref{tab:pt2}. The ORLs for Problematic Trajectory 2 are Green Park, with the following durations (in hours):\begin{quotation}\noindent12.92; 119.92; 23.75; 12.42; 3.25; 4; 8.83; 71.94; 24; 10.17; 9.67\end{quotation} Stockwell (with duration 11.08) and Clapham North (with duration 11.25). The pre-processing has labelled the most likely home station as Stockwell Station and the most likely work station as Green Park. In contrast, visual inspection suggests that this person commutes from Clapham North (home) to Green Park (work). In this instance Stockwell is an intermediate stop (very close to Clapham North) where the passenger changes train. For some reason wireless connections appear to be established more frequently in Stockwell that in Clapham North and thus the mis-labelling is understandable in this instance.

\begin{table}
\begin{center}
\begin{tabular}{p{5cm} P{2cm}}
\small
Journey & Count \\
\hline
\hline
Green Park & AM (15), PM (0) \\
\hline
Balham & AM (7), PM (0) \\
\hline
Stockwell & AM (0), PM (6) \\
\hline
Vauxhall Station & AM (0), PM (4) \\
\hline
Clapham North & AM (0), PM (4) \\
\hline
Oxford Circus & AM (3), PM (0) \\
\hline
Stockwell \(\rightarrow\) Green Park & AM (3), PM (0) \\
\hline
Victoria \(\rightarrow\) Green Park & AM (0), PM (3) \\
\hline
Vauxhall Station \(\rightarrow\) Stockwell & AM (0), PM (2) \\
\hline
Notting Hill Gate Station & AM (0), PM (2) \\
\hline
Clapham South & AM (2), PM (0) \\
\hline
Green Park \(\rightarrow\) Stockwell & AM (0), PM (2) \\
\hline
Baker Street & AM (2), PM (0) \\
\hline
Notting Hill Gate Station \(\rightarrow\) Parsons Green Station \(\rightarrow\) Victoria & AM (0), PM (1) \\
\hline
Clapham North \(\rightarrow\) Victoria \(\rightarrow\) Baker Street & AM (1), PM (0) \\
\hline
Kennington \(\rightarrow\) Stockwell & AM (0), PM (1) \\
\hline
Clapham South \(\rightarrow\) Stockwell \(\rightarrow\) Pimlico Station \(\rightarrow\) Green Park & AM (1), PM (0) \\
\hline
Green Park \(\rightarrow\) Victoria & AM (0), PM (1) \\
\hline
Green Park \(\rightarrow\) Vauxhall Station \(\rightarrow\) Clapham North & AM (0), PM (1) \\
\hline
Victoria \(\rightarrow\) Clapham North & AM (0), PM (1) \\
\hline
...
\end{tabular} 
\end{center}
\caption{The condensed journey list of Problematic Trajectory 2. The supplement \cite[Section 2.2]{karlsen-2018-supplement} displays the full pre-processing output for this trajectory.}
\label{tab:pt2}
\end{table}


Problematic Trajectory 3 shows an example of reverse labelling. The journey string is as follows: \begin{quotation}\noindent\seqsplit{U|O|UOH|U|OUUUUOH|WOUO|UUUO|UUW|WOUUO|WOU|WOUH|H|UUUO|W|WOUUUO|UUO|OUUO|U|OUW|OUUOH|UUUOW|UO|O|H|WHO|U|H|OU|UUH|OUOUUUUU|WUUO|W|OUUO|OUUUOH|UUO|UO|W|W|WOU|OUUOUO|W|WO|W|OUUOHOUUOW|OH|UUO|UUO|OUUUUUO|OUUO|WOUUUO|OUUOH|OO|UU|WUUOUUOW|OUUOH|OUUO|UU|WOOH|UUU|WOUUUUOH|OUU|OUUOH|OW|W|W|OUUUOW|U|UUOH|UUUU|UO|U|W|WOUUO|OOW|WOUUUOH|H|OUOH|OUOH|WOUO|OUO|OUU|OUUUOW|WOUUO}\end{quotation} The abbreviated form of the output from the preprocessing is shown in Table \ref{tab:pt3}. The ORLs for Problematic Trajectory 3 are Arnos Grove Station, with the following durations (in hours):\begin{quotation}\noindent17.28; 28.85; 173.4; 21.88; 9.2; 11.17\end{quotation} Bounds Green (with durations 18.41; 2.31; 11.35), Kings Cross \& St Pancras with durations:\begin{quotation}\noindent11.21; 12.24; 8.4; 8.2\end{quotation} and the Barbican Station (with duration 43.58). The pre-processing has labelled the most likely home station as Barbican Station Station and the most likely work station as Arnos Grove Station. However, visual inspection suggests that the commuter commutes from a home location of Arnos Grove to the Barbican station, with a change at King's Cross St. Pancras. A possible explanation for this is the unusual times of some of the commutes. For example, the commuter leaves Arnos Grove one Friday at 11:26 which is outside of the 05:00--10:00 time bracket where the ORL receives extra weighting (see Section \ref{sect:trl} for details of this weighting mechanism). This is not an unusual occurrence for this commuter, meaning that the work and home labelling is ineffective in this instance.
	
\begin{table}
\begin{center}
\begin{tabular}{p{5cm} P{2cm}}
\small
Journey & Count \\
\hline
\hline
Arnos Grove Station & AM (0), PM (9) \\
\hline
Barbican Station & AM (0), PM (4) \\
\hline
Bounds Green \(\rightarrow\) Wood Green \(\rightarrow\) Turnpike Lane \(\rightarrow\) Kings Cross and St Pancras \(\rightarrow\) Barbican Station & AM (0), PM (3) \\
\hline
Wood Green \(\rightarrow\) Turnpike Lane \(\rightarrow\) Highbury and Islington Station \(\rightarrow\) Kings Cross and St Pancras & AM (2), PM (0) \\
\hline
Holloway Road Station \(\rightarrow\) Turnpike Lane \(\rightarrow\) Bounds Green & AM (0), PM (2) \\
\hline
Wood Green & AM (0), PM (2) \\
\hline
Arnos Grove Station \(\rightarrow\) Bounds Green \(\rightarrow\) Wood Green & AM (2), PM (0) \\
\hline
Turnpike Lane \(\rightarrow\) Wood Green \(\rightarrow\) Bounds Green & AM (0), PM (2) \\
\hline
Arnos Grove Station \(\rightarrow\) Wood Green \(\rightarrow\) Turnpike Lane \(\rightarrow\) Kings Cross and St Pancras \(\rightarrow\) Holloway Road Station \(\rightarrow\) Wood Green \(\rightarrow\) Bounds Green \(\rightarrow\) Arnos Grove Station & AM (1), PM (0) \\
\hline
Kings Cross and St Pancras \(\rightarrow\) Holloway Road Station \(\rightarrow\) Wood Green \(\rightarrow\) Bounds Green & AM (1), PM (0) \\
\hline
Kings Cross and St Pancras \(\rightarrow\) Holloway Road Station \(\rightarrow\) Arsenal Station \(\rightarrow\) Bounds Green & AM (1), PM (0) \\
\hline
Bounds Green & AM (0), PM (1) \\
\hline
Wood Green \(\rightarrow\) Turnpike Lane \(\rightarrow\) Barbican Station & AM (1), PM (0) \\
\hline
St Paul's Station & AM (0), PM (1) \\
\hline
...
\end{tabular} 
\end{center}
\caption{The condensed journey list of Problematic Trajectory 3. The supplement \cite[Section 2.3]{karlsen-2018-supplement} displays the full pre-processing output for this trajectory.}
\label{tab:pt3}
\end{table}


Problematic Trajectory 4 contains an example trajectory that contains insufficient information. The journey string is as follows: \begin{quotation}\noindent\seqsplit{U|UUUU|UUUUU}\end{quotation} The abbreviated form of the output from the preprocessing is shown in Table \ref{tab:pt4}. There were no ORLs identified for Problematic Trajectory 4. For this reason the pre-processing has been unable to label the most likely home station and the most likely work station. It is essentially impossible to analyse such trajectories because none of the locations repeat sufficiently for the ORL extraction and labelling code to work. Many of the instances in the database are of this type. 

\begin{table}
\begin{center}
\begin{tabular}{p{5cm} P{2cm}}
\small
Journey & Count \\
\hline
\hline
Canning Town Station & AM (0), PM (1) \\
\hline
Earls Court Station \(\rightarrow\) Boston Manor Station \(\rightarrow\) Heathrow Terminal 4 \(\rightarrow\) Heathrow Terminals 1,2, and 3 & AM (0), PM (1) \\
\hline
Acton Town \(\rightarrow\) Hounslow East Station \(\rightarrow\) Hounslow West Station \(\rightarrow\) Heathrow Terminal 4 \(\rightarrow\) Heathrow Terminals 1,2, and 3 & AM (1), PM (0) \\
\end{tabular} 
\end{center}
\caption{The condensed journey list of Problematic Trajectory 4. The supplement \cite[Section 2.4]{karlsen-2018-supplement} displays the full pre-processing output for this trajectory.}
\label{tab:pt4}
\end{table}


Problematic Trajectory 5 remains unlabelled even though 21 journeys exist. The journey string is as follows: \begin{quotation}\noindent\seqsplit{UUUUU|UUU|UUUUUU|UUU|UUUUUUUU|UUUUUUU|UUUUUUUU|UUUUUU|UU|UUUUU|U|UUUUUU|U|UUUUUUUUU|U|UUUUUU|UUUUUUU|UUUUUUUU|UUUUUUUU|UUUUUUU|UUUUUUUU}\end{quotation} The abbreviated form of the output from the preprocessing is shown in Table \ref{tab:pt5}. Again, no ORLs have been identified for the journeys. Since the code that labels locations and work or home depends upon there being a list of offline rest locations to choose from, the code that labels offline rest locations as work or home cannot function, thus the locations remain unlabelled.

\begin{table}
\begin{center}
\begin{tabular}{p{5cm} P{2cm}}
\small
Journey & Count \\
\hline
\hline
Charing Cross & AM (0), PM (2) \\
\hline
London Bridge Station \(\rightarrow\) Southwark Station \(\rightarrow\) Westminister Station \(\rightarrow\) Green Park \(\rightarrow\) Marble Arch \(\rightarrow\) Lancaster Gate \(\rightarrow\) Queensway Station \(\rightarrow\) Shepherd's Bush & AM (2), PM (0) \\
\hline
London Bridge Station \(\rightarrow\) Green Park \(\rightarrow\) Marble Arch \(\rightarrow\) Lancaster Gate \(\rightarrow\) Queensway Station \(\rightarrow\) Notting Hill Gate Station \(\rightarrow\) Shepherd's Bush & AM (1), PM (0) \\
\hline
London Bridge Station \(\rightarrow\) Southwark Station \(\rightarrow\) Westminister Station \(\rightarrow\) Green Park \(\rightarrow\) Marble Arch \(\rightarrow\) Lancaster Gate \(\rightarrow\) Notting Hill Gate Station \(\rightarrow\) Shepherd's Bush & AM (1), PM (0) \\
\hline
London Bridge Station \(\rightarrow\) Southwark Station \(\rightarrow\) Westminister Station \(\rightarrow\) Lancaster Gate \(\rightarrow\) Notting Hill Gate Station \(\rightarrow\) Shepherd's Bush & AM (1), PM (0) \\
\hline
Notting Hill Gate Station \(\rightarrow\) Queensway Station \(\rightarrow\) Charing Cross & AM (0), PM (1) \\
\hline
London Bridge Station \(\rightarrow\) Southwark Station \(\rightarrow\) Green Park \(\rightarrow\) Lancaster Gate \(\rightarrow\) Queensway Station \(\rightarrow\) Shepherd's Bush & AM (1), PM (0) \\
\hline
Queensway Station \(\rightarrow\) Lancaster Gate \(\rightarrow\) Marble Arch & AM (0), PM (1) \\
\hline
Southwark Station \(\rightarrow\) Westminister Station \(\rightarrow\) Green Park \(\rightarrow\) Queensway Station \(\rightarrow\) Shepherd's Bush & AM (1), PM (0) \\
\hline
...
\end{tabular} 
\end{center}
\caption{The condensed journey list of Problematic Trajectory 5. The supplement \cite[Section 2.5]{karlsen-2018-supplement} displays the full pre-processing output for this trajectory.}
\label{tab:pt5}
\end{table}


In summary, manual inspection reveals that the pre-processing has worked correctly on a number of trajectories. However, this is not always the case. Some trajectories are simply too small to analyse correctly. Some trajectories are mis-labelled because one or more locations at the end points of the journey do not always show up in customer trajectories (possibly due to the reasons identified within Section \ref{sect:daintro}). Some trajectories are reverse-labelled due to strange commute times. Finally, some trajectories remain unlabelled because no offline rest locations have been identified. These limitations represent areas for improvement that could be addressed through future work (see Section \ref{sect:fwork}).

\subsection{Automated Checking of Pre-Processed Output}

For this work a number of ``domain-specific'' tests were coded to test that the pre-processing steps function correctly on a number of test trajectories according to what we know about these trajectories from manual inspection. 14 test cases where selected and manually analysed (including the five example commutes above). For these 14 test cases, a number of sets of tests were created to evaluate the pre-processing. There are currently 14 such tests for location tagging (where each test checks that both home and work locations have been tagged correctly) and 14 equivalent tests for the overall trajectory pattern. (The journey pattern test is used to verify that the trajectory has been split up in to journeys in the correct way -- if the splitting was incorrect it is extremely likely that the resultant journey test would be substantially different from that expected.)

The results are as follows. Thirteen of the 14 trajectory pattern tests pass and 13 of the 14 home and work labelling tests pass. The one that fails does so due to a mis-label. The expected home station is ``Canada Water Station" whilst the expected work station is ``Canary Wharf". The home and work stations have been reversed and thus the home and work label tests fail and the trajectory pattern is incorrect. Correction of this issue would require further work as suggested within Section \ref{sect:fwork}.

\section{Data Visualisation and Clustering}

\subsection{Features Selected For Use}

Of the original 11 potential features listed within Section \ref{sect:ppint}, five are selected for use for the data visualisation and clustering:

\begin{itemize}
\item average journey duration
\item the journey pattern (work to home, home to work, ...)
\item number of locations visited per journey
\item journey frequency
\item number of journeys per `offline rest location'
\end{itemize}
The remaining features were discounted for various reasons. Total time spent travelling is dependant upon how long the user has been using the service. Since the length of time that the user has been using the service is not relevant to distinguishing between consumer types this feature was discarded. Total distance of travel (direct or the distance along the route actually travelled) was not implemented for technical reasons -- it requires detailed knowledge of each route that we do not currently possess. Distance of each journey and distance travelled per unit of time were also not implemented for this reason. The times of day person is travelling has become part of the journey pattern based on the rest location labelling described in Section \ref{sect:trl} and is thus not included separately. The phone type of the user was not included due to the complexity of constructing a distance measure that could be applied to phone types. Phone types are discussed further in Section \ref{sect:fwork}.

\subsection{Basic Data Visualisation}
\label{sect:scatter}

\begin{figure*}
\begin{tabular}{cc}
\fbox{\includegraphics[width=\columnwidth]{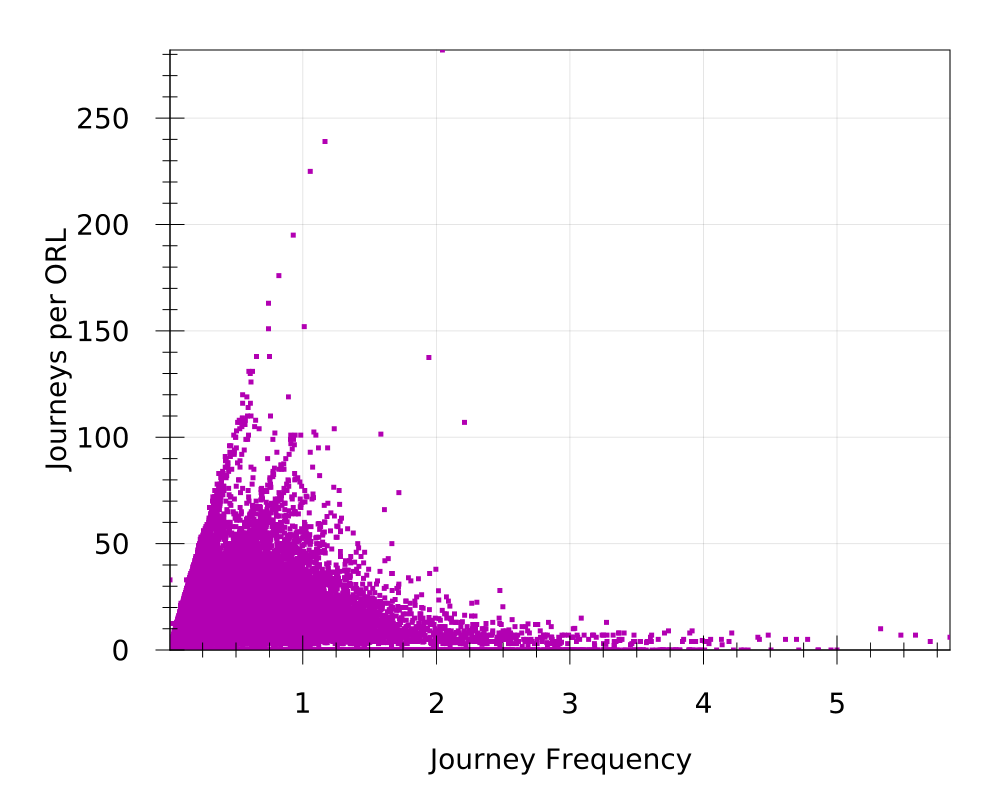}} & \fbox{\includegraphics[width=\columnwidth]{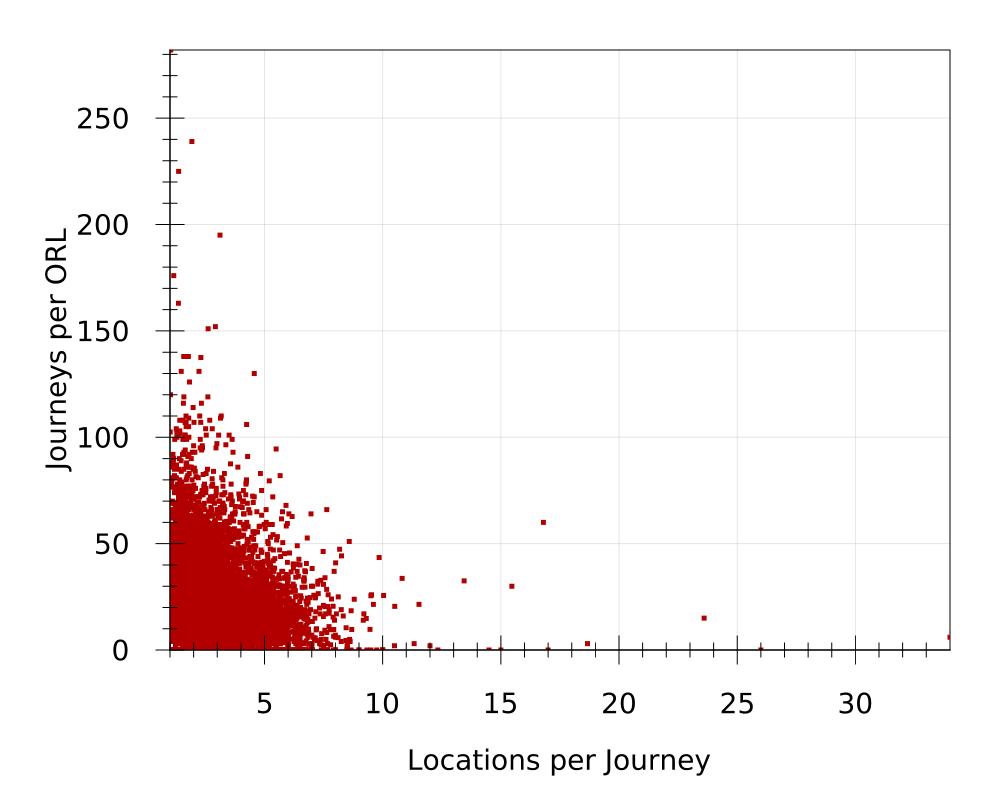}} \\ 
\fbox{\includegraphics[width=\columnwidth]{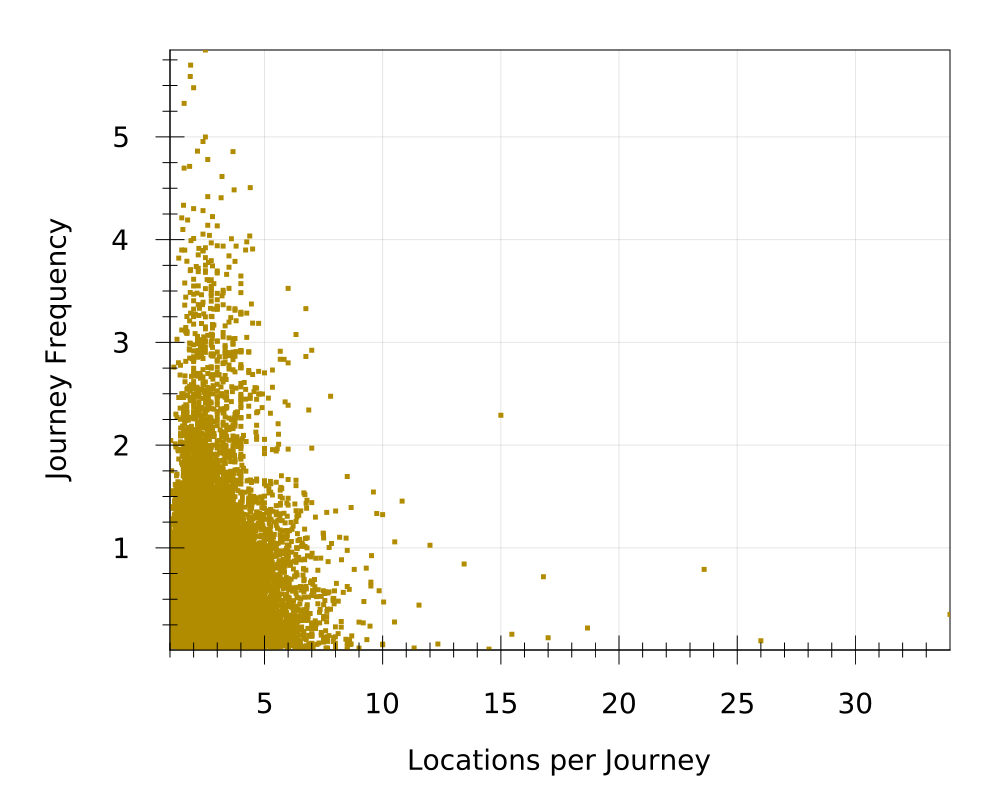}} & \fbox{\includegraphics[width=\columnwidth]{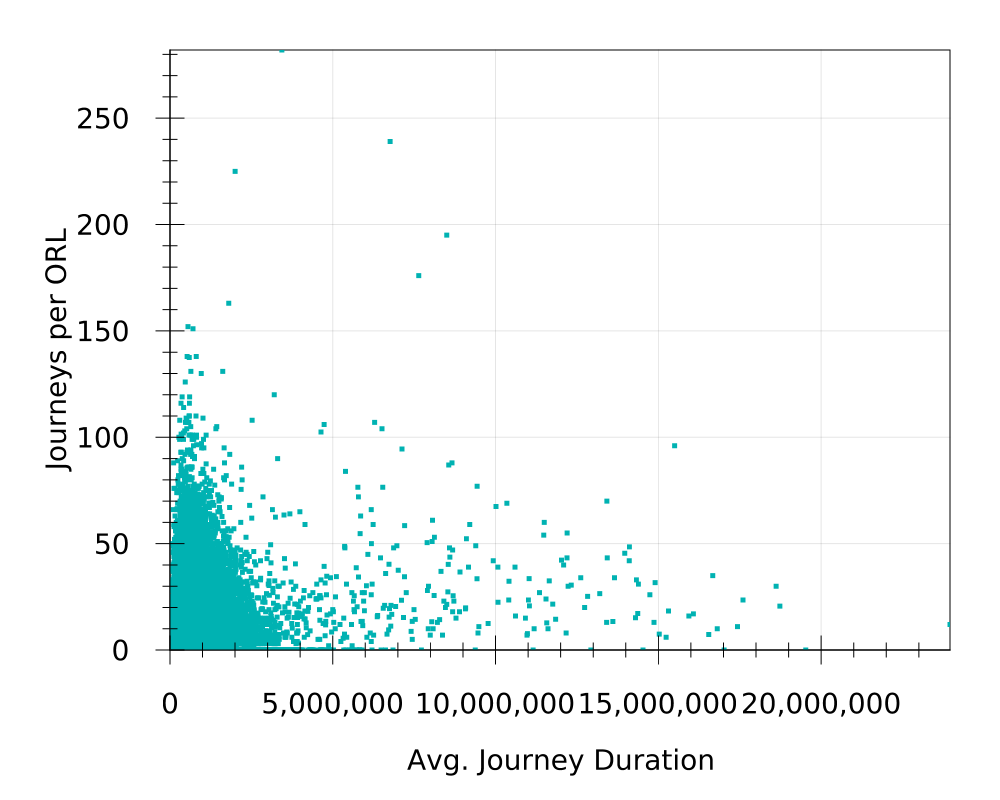}} \\ 
\fbox{\includegraphics[width=\columnwidth]{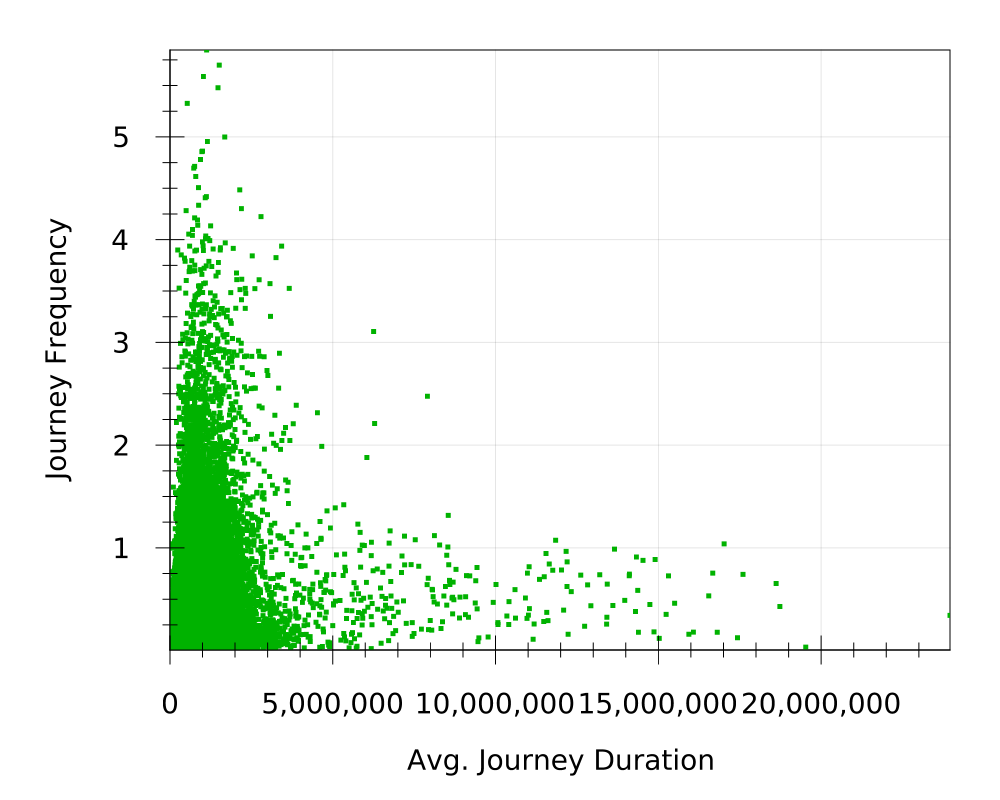}} & \fbox{\includegraphics[width=\columnwidth]{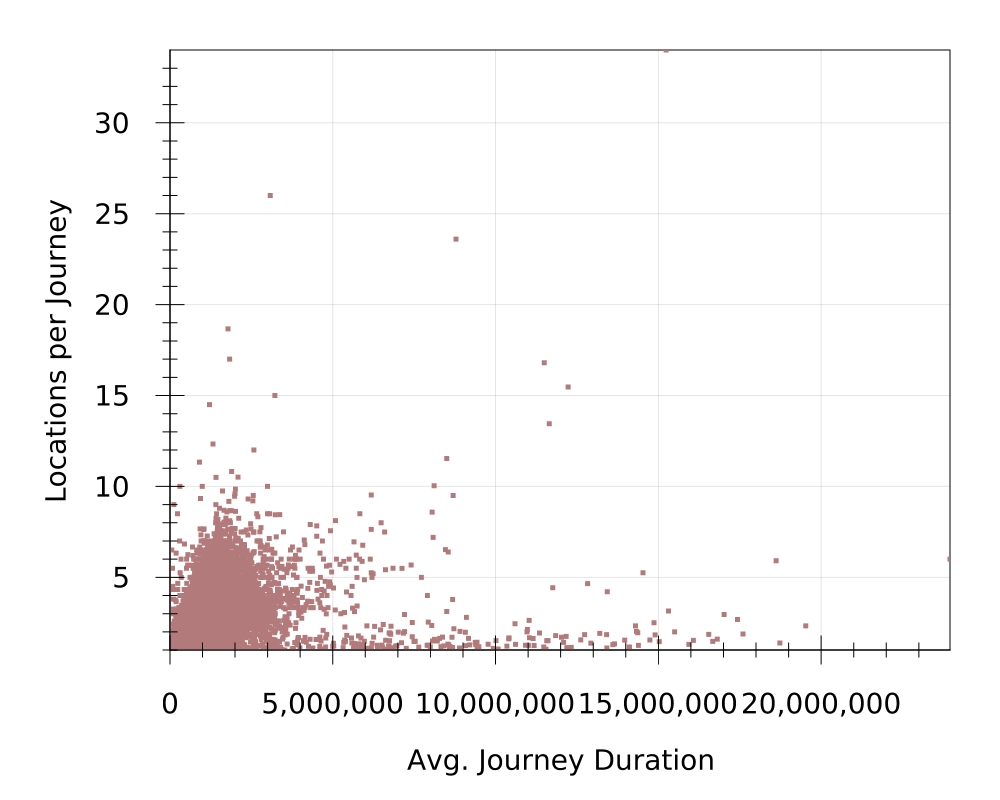}} \\ 
\end{tabular}
\caption{scatter-plots for the four numerical variables used in the clustering. The scatter-plots are from every trajectory with \(\geq 10\) locations that have trajectory points recorded over 24 hours or more (98,601 trajectories in total).}
\label{fig:sp1}
\end{figure*}

A simple approach to identifying clusters in data is through visual inspection. This may be done either via a scatter-plot matrix (where each pair of variables is plotted in their own small chart, and all of the charts are shown within a matrix to enable easy `visual scanning' by an observer) or via some kind of dimensionality reduction. Scatter-plots have been prepared for each variable against every other variable used in the clustering. Since one property used in the clustering is non-numerical, except when later converted to a numerical distance measure when two items are compared, there are four variables to produce scatter-plots for. The 6 relevant plots are produced and shown within Figure \ref{fig:sp1}.

An examination of the scatter-plots within Figure \ref{fig:sp1} suggests that the data consists of a single large cluster with some outliers. We can see that in terms of journeys per ORL, the majority of people have a value of less than 75. Some consumers (those with very few ORLs) have a large journey per ORL value indicating that they consistently travel to the same location. The journey frequency has a similar pattern, with the majority of customers travelling less than or equal to twice per day, with a far smaller number of people travelling more than twice per day. Locations per journey is consistently less than 8 with only a few outliers with a greater value than this. We can also see that average journey duration is almost always less than 4,000,000 milliseconds (around 66 minutes), again with some outliers. Two patterns of interest can be seen in the scatter-plots. Firstly, when charting journey frequency against average journey duration we can see that travellers tend to either travel for long periods of time or travel frequently -- there are very few travellers that take a moderate value on both variables. Secondly, we can see there tends to be a positive relationship between value of journeys per ORL and journey frequency. This is due to the relationship between the number of days the passenger has been using the service and the number of offline rest locations.

\subsection{Dimensionality Reduction}
\label{sect:drdv}

\paragraph{Description.} Dimensionality reduction approaches reduce the number of dimensions in the data such that the data can be charted for visual inspection. Principal Component Analysis \cite{abdi-2010-principal} is one approach to dimensionality reduction. PCA performs a kind of `lossy compresion' on the features of the data set \cite{bro-2014-principal}, resulting in a new set of extracted features ordered by their explanatory power. Whilst PCA can produce any number of extracted features less than the original number of feature, more of the variance in the data is explained by the first 2-3 new features \cite{bro-2014-principal}. Thus, for example, eight features transformed using PCA will result in fewer than 8 Eigenvectors (new extracted features). Each successive Eigenvector explains progressively less of the variance of the data. The first two or three Eigenvectors thus explain the majority of the variance and can be charted to provide an informative look at the data. Visual inspection can then be used to look for any clusters present. 

\paragraph{Application.} Here we apply PCA to the 10,000 customers selected for use in the data clustering (see Section \ref{sect:sample}).\footnote{We use the PCA implementation from the Apache Commons Math Library \cite{apache-2018-math}.} The data was normalised before PCA was applied. PCA was selected over other dimensionality reduction approaches due to its established nature and ubiquity. It should be noted that \emph{here PCA is primarily used as a means to obtain a visualisation for the cluster analysis} performed in the proceeding section.

\subsection{Data Clustering}
\label{sect:dc}

\subsubsection{Distance Measures}

Clustering algorithms usually make use of a distance measure for comparing individual items within the data set such that a distance can be worked out between every pair in the data set. This distance measure is then used by the clustering algorithm to group together similar items and assign different items to distinct groups.

Distance measures used in clustering are often simple (such as Euclidean distance \cite{everitt-2001-cluster} or City Block/Manhattan distance \cite{everitt-2001-cluster}). However, there are also more specialised distance measures available, such as the Levenshtein distance, used to compare strings of characters.\footnote{Levenshtein distance was used in an earlier version of the distance measure described below but it is not longer used due to a built-in bias towards short journey patterns.}

\subsubsection{Clustering Algorithm Selection}

Selecting a clustering algorithm is a non-trivial task and no precise algorithmic approach to doing so yet exists. For our present dataset, hierarchical clustering algorithms are deemed unsuitable first of all. Hierarchical algorithms require a distance matrix be created that includes the distance between every pair of items in the data set. Since there are 348,304 items in our data set at present, working out this distance matrix would require 121,315,328,112 distance calculations. Furthermore, the resulting dendrogram produced when there are a great many items to cluster is often incomprehensible because the labels of the diagram's leaves are so small or overlap. Whilst alternative visualisation approaches exist \cite{bisson-2012-improving}, other more appropriate solutions may be preferable.

We also rule out the popular K-means algorithm here for two reasons. Firstly, the K-means algorithm requires that the user specify the number of clusters beforehand \cite{morissette-2013-k}. Since we do not know here how many clusters are present in the data set, this is a problem. Furthermore, the K-means algorithm can only identify spherical clusters effectively \cite{jain-2010-data}. Non-spherical clusters tend to be partitioned in to multiple clusters instead, producing a poor quality solution.




Consideration of density-based solutions suggest that these algorithms do not have the above issues. One does not have to specify the number of clusters (though some parameters do remain to be specified), an exhaustive distance matrix is not required, and they can cope with many data points. Here we use DBSCAN \cite{ester-1996-density}, described in more detail below. Though the algorithm has some limitations (it has difficulty identifying some clusters when the dataset has clusters of differing density) it provides a good first choice with which to cluster.

\subsubsection{DBSCAN}

\paragraph{Description.} DBSCAN \cite{ester-1996-density} is tried and tested (the original paper has been cited more than 11,000 times on Google Scholar). Furthermore, DBSCAN does not insist that all data items belong to one or other of the clusters. Whilst DBSCAN attempts to add unprocessed points to one of the clusters, if the conditions are not met then the points will be marked as noise instead. This feature improves the overall quality of the clustering.

The DBSCAN algorithm has two parameter settings: `\emph{MinPts}' and `\emph{Eps}'. \emph{Eps} is the radius with which to search when searching for the neighbours of a current point \cite{li-2018-smile}. \emph{MinPts} is the minumum number of points required within \emph{Eps} distance of the current point (\emph{including the current point}) for a point to become a member of a cluster (as opposed to becoming an outlier) \cite{ester-1996-density}. For the clustering runs here, \emph{MinPts} was set to 10 and \emph{Eps} was set to 0.04.

The algorithm functions as follows. An initial point is selected. The number of points within radius \emph{Eps} of that point are then counted. If the number of points within radius \emph{Eps} (including the central point) is greater than or equal to \emph{MinPts} then those points become a member of a cluster. The process is then repeated with the neighbouring points discovered during this first step (so long as the \emph{MinPts} test was passed). This is repeated until the `mining' of the cluster is exhausted. If there are insufficient points within \emph{Eps} and the object is not part of a cluster as a neighbouring point, it is marked as noise. Once one cluster is exhausted the process re-starts on the next unlabelled point. This is repeated until all points have been labelled.

\paragraph{Implementation.} The DBSCAN algorithm is present in a number of clustering and/or machine learning libraries such as Python's scikit-learn \cite{scikit-2018-clustering}, the Java-based Apache Commons Math library \cite{apache-2018-math} and the Smile machine learning framework \cite{li-2018-smile} -- also Java-based. The Apache commons library is of limited help here due to the apparent inability to specify one's own distance measure. However, the Smile framework appears suitable for our needs. The algorithm is present, the distance measure can be overridden, and the framework appears to be relatively mature. In order to test the implementation a stand-alone implementation of DBSCAN will also be created to verify that the Smile DBSCAN algorithm works as intended.\footnote{The implementation of a stand-alone DBSCAN implementation, combined with testing, revealed that, whilst Smile adheres reasonably closely to the original DBSCAN specification, the \emph{MinPts} parameter in DBSCAN does not include the current central point itself. This is contrary to the original DBSCAN specification so the stand-alone implementation was used to produce the results below.}

\subsubsection{Distance Measure Used}

Here we use a distance measure that is the weighted sum of five different components:

\begin{itemize}
\item (\(d_1\)) -- the distance between the consumer patterns
\item (\(d_2\)) -- the difference between the journey frequencies
\item (\(d_3\)) -- the difference between the locations per journey
\item (\(d_4\)) -- the difference between the average journey durations
\item (\(d_5\)) -- the difference between the number of journeys per offline rest location
\end{itemize}

The values used for the components \(d_2\) to \(d_5\) are scaled to the range 0 to 1 before their differences are calculated. The component \(d_1\) is scaled to the range 0 to 1 after the distance between the consumer patterns is calculated since, due to the nature of the journey patterns, no numerical representation of an \emph{individual} consumer pattern is available. The individual components are then summed and weighted. The weight applied to each component is 0.20. The total distance between any two points is thus:

\[ d(p_1,p_2) = \frac{1}{N}\sum_{i=1}^{N} d_i \] where \(N=5\) in this instance.

The distance between the consumer patterns is calculated using the previously constructed map that each consumer object holds (see Section \ref{sect:jpatt}), relating each unique journey pattern that the consumer has with a count of that journey pattern. When the consumer pattern distance is desired, a list of patterns related just to consumer 1 is prepared, as is a list of patterns related just to consumer 2. The remaining patterns are assigned to a list of shared patterns. The total distance is the count of the number of instances of each unique pattern related to just consumer 1 and just consumer 2, added to half the value of the difference between the counts for each pattern shared by the two consumers. (Thus if one pattern is shared and consumer 1 has used this pattern 4 times and consumer 2 has used it 8 times then the distance is increased by 2.)\footnote{Caveat: whilst the approach above works significantly better than the earlier attempt at using the Levenshtein distance measure here, the above approach does have a limitation. It does not currently recognise the closeness of two different consumer patterns. Again annotating locations in the journey pattern as (H)home, (W)ork or (U)nknown, one pattern of \texttt{H,W} and one pattern of \texttt{H,U,W} are not considered any closer than one pattern of \texttt{H,W} and one of \texttt{U,U}. A possible future improvement to the distance measure is described within Section \ref{sect:predict}.}

\subsubsection{Sampling and Parameter Settings}
\label{sect:sample}

At present a subset of the full data is used due to memory constraints. In the last run performed, 10,000 customers were randomly selected and used for the clustering process. The minimum number of locations in a valid trajectory was set to 10. This means that, of the subset of customers with a trajectory of at least 10 locations, 10,000 were selected randomly and used in the clustering process. It should be noted that though 10,000 customers were used, 74,165 trajectories were first discarded due to the filters on trajectory length, time and values.

The parameter settings used for the data clustering run are shown in Table \ref{tab:params}.\footnote{The EARLIEST\_VALID\_TIMESTAMP is specified as the start of the year 2000 in Unix Time -- milliseconds since 1970-01-01 00:00.}

\begin{table*}
\small
\begin{center}
\begin{tabular}{|c|c|c|}
\hline 
Parameter & Setting & Unit \\ 
\hline 
MinPts & 10 & Points \\ 
\hline 
Eps & 0.04 & N/A \\ 
\hline 
MAX\_TIME\_BETWEEN\_POINTS\_IN\_JOURNEY & 1000 x 60 x 80 & Milliseconds \\ 
\hline 
MIN\_TIME\_FOR\_ORL & 1000 x 60 x 30 & Milliseconds \\ 
\hline 
MIN\_TRAJECTORY\_LENGTH & 10 & Locations \\ 
\hline 
MAX\_NUM\_TRAJECTORIES & 10,000 & Trajectories \\ 
\hline 
EARLIEST\_VALID\_TIMESTAMP & 946684800 & Milliseconds \\ 
\hline 
MIN\_NUM\_DAYS\_DATA\_FOR\_VALID\_TRAJ & 1 & Days \\ 
\hline 
\end{tabular} 
\end{center}
\caption{The parameter settings used for the data clustering run.}
\label{tab:params}
\end{table*}

\subsubsection{Results}
\label{sect:results}

\begin{figure}
\fbox{\includegraphics[width=0.99\linewidth]{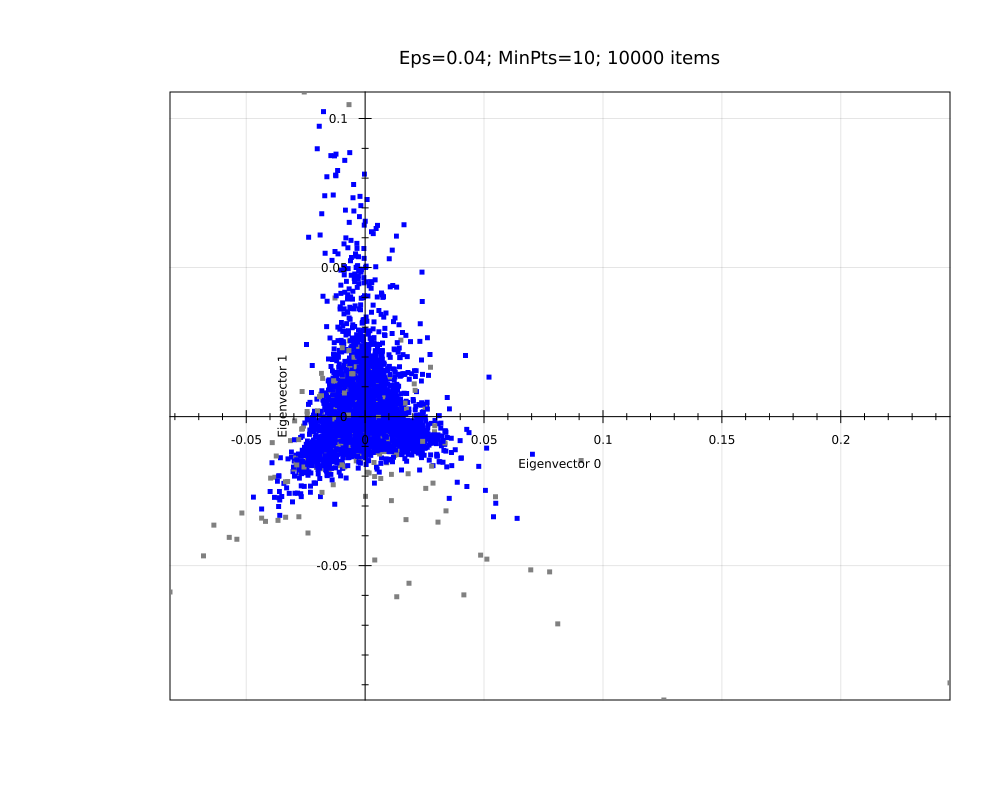}}
\caption{Hybrid output from the Principal Component Analysis and DBSCAN-based clustering. Points deemed to be noise are coloured grey, whilst each cluster identified is assigned its own colour. Present results indicate that only a single cluster is present within the data.}
\label{fig:res1}
\end{figure}

Presently a number of runs have been performed with a range of parameter settings for the DBSCAN algorithm. The results presented here are for the parameter settings \emph{MinPts} set to 10 and \emph{Eps} set to 0.04. It has been found that, despite much adjustment of the parameter settings, a single large cluster is produced, with some peripheral outliers and noise. See Figure \ref{fig:res1} for details. The scatter-plots for the four numerical variables shown in Figure \ref{fig:sp1} also support this conclusion.

\section{Future Work}
\label{sect:fwork}

\subsection{Integration of CCTool}

CCTool \cite{moschoyiannis-2016-web-based} is a software tool designed to identify the most suitable intervention points within complex systems. The input for the tool consists of a network model constructed through participation with stakeholder groups. The output consists of sets of proposed `lever points' for the modelled system.

The network model is constructed of nodes and links, with associated information attached to both the nodes and links. Each node (graph vertex) is associated with an indicator of intervention difficulty at that node. Each directed link (graph edge) is associated with an indicator of the strength of the relationship between the node that the directed link originates from and the node that the directed link is connected to.

The network model, constructed within the tool\footnote{Available at \url{http://cctool.herokuapp.com/}}, is analysed to provide a list of all the `minimum control configurations' available \cite{moschoyiannis-2016-web-based}. (Each minimum control configuration consists of a number of nodes that, when used together, are able to move the system from any one state to any other state within a finite amount of time.) 

It is possible that, as part of the future work on the OJPA project, this tool could be adapted for use within the ``Recommendation Rule Engine''.

\subsection{Learning Classifier Systems}

It is possible that Learning Classifier Systems (a machine learning approach involving genetic algorithms) \cite{urbanowicz-2009-learning} could be used as part of the Recommendation Engine. The rule structure of a Learning Classifier System is well suited to store the relationship between particular customer preferences (and current environmental factors) and associated suggestions provided to the user. In particular, I suggest an integer-valued \cite{wilson-2000-mining} or real-valued \cite{wilson-1999-get} XCS implementation.

XCS rules are in a condition:action format. An example rule might look as follows:

\begin{quotation}\footnotesize\texttt{[4,5][1,2][3,5][5,5][4,5] : 42}\end{quotation}

In this simple rule we have five properties and one action (42). If match conditions on the properties are met then the action is triggered. For the above rule to trigger, the value of property one would have to be a 4 or 5, the value of property two would have to be a 1 or 2, the value of property 3 would have to be in the range 3 to 5, the value of the fourth property would have to be a 5 and the value of the fifth property would have to be a 4 or 5. An example of using such a rule would be as follows. Property one could be speed preference, property two could be cost preference, property three could be comfort and productivity, property four could be bad weather preference and property five could be the current state of the weather. 42 could be `suggest individual taxi booking'. 

An example of a rule match is now provided. `James' has a strong preference for getting where he wants to go quickly so rates speed as 5 via the chat bot. He rates cost as a 2. He rates comfort and productivity as a 4. He rates `bad weather avoidance' at 5 (he wears expensive suits that he does not want to get wet). When James is travelling, the current weather is a 4 -- quite heavy rain. The OJPA mobile app uses a weather API to obtain the `4' for weather, combines this with James' input preferences (5,2,4,5) and the OJPA app calls the recommendation engine (located on a server) via a REST call. The input string 52454 is fed in. This matches with the rule \texttt{[4,5][1,2][3,5][5,5][4,5] : 42} and thus the action 42 is triggered. The action 42 is located within the action table and identified as `suggest individual taxi booking' and thus the OJPA app suggests the individual taxi booking.

The above presents a small example of the use of a single rule. Learning Classifier Systems hold many such rules and constantly evolve new ones, whilst removing old unsuccessful rules and retaining old but successful rules \cite{urbanowicz-2009-learning}. (The relevance of suggestions provided to users via the OJPA application can be used as feedback to the LCS to assist in the LCS learning and evolving better rules.) In this way, a Learning Classifier System would be able to hold all of the knowledge required to provide customised suggestions to customers in a context-specific manner.

\subsection{Possible Clustering Improvements}

There are several possible improvements and future work ideas that can be pursued. The 80-minute-gap rule for breaking up the trajectory in to journeys is somewhat arbitrary. Some journeys may take more than 80 minutes and be split in to two or more separate journeys erroneously. For this reason it is possible that we should consider actual travel time between points when working out where to break up a trajectory. The splits could then be made where the time between the points is greater than the estimated journey time plus some margin of error (say 30 minutes). Estimated journey time plus 30 minutes is likely to be far more accurate than a fixed 80 minutes. To take action on this we would either have to acquire or build a weighted graph of the railway network where the weights signify journey times between locations.

One alternative distance measure for journey patterns suggests itself from the traditional clustering literature. One measure of distance between two clusters is average linkage \cite{everitt-2001-cluster}. This is the average distance between all points in cluster A and all points in cluster B. For example, where there are 3 points in cluster A and 2 in cluster B, the distance is \(d(c_A,c_B) = (d_{1,1} + d_{1,2} + d_{2,1} + d_{2,2} + d_{3,1} + d_{3,2}) / 6\) \cite{everitt-2001-cluster}. This distance measure could be adapted for use measuring the distance between two trajectories. The trajectories could be split in to journeys and then the distance measure could be applied to measure the distance between the two sets of journeys (using LCSS to measure the distance between any two given journeys). This could potentially be superior to pure LCSS, because it incorporates distinct journeys and takes journey frequency in to consideration, though it has the probable down side of requiring greater effort to compute.

A database of phones could be constructed and then used to construct a dissimilarity measure for the phone details such that we can get an understanding of how close or distant any two customers are in terms of the mobile phones they use. This distance measure could then be used as part of the overall distance measure, possibly leading to greater differentiation between customers.

It would also be possible to improve the code that labels work and home by altering the requirement that work and home be offline rest locations. It is possible that the wireless data is such that no ORLs are identified. Under these circumstances work and home remain unlabelled. There is also the unlikely, but potentially problematic case, of a number of locations being labelled as ORLs without work and home being labelled as ORLs. In this situation it is possible that the wrong ORLs are labelled as work and home. To fix this it could be possible to change the way that work and home are labelled to consider locations not annotated as an ORL.

\subsection{Route Prediction and Train Identification}
\label{sect:predict}

In the context of the OJPA project it will be useful to predict the route of each customer in order to forecast the ultimate destination and intermediate waypoints in the customer's journeys. These predictions will be required in order to inform customers of cancelled trains or suggest onward journey options without requesting that the user input their destination for every journey. In this context it is useful to identify routes that are \emph{geographically} similar (i.e. trajectory clustering rather than semantic trajectory clustering).

A simple approach to identifying geographically similar trajectories involves applying a Longest Common Substring algorithm \cite{200success-2018-finding} to each pair of journey patterns. This report's supplemental paper \cite[Section 3]{karlsen-2018-supplement} contains a target trajectory, with 5 similar trajectories (identified using a longest common substring algorithm -- i.e. those deemed most similar to the target trajectory are those that share the longest common sequence of locations). However, whilst the final two trajectories share many locations with the target trajectory the majority of journeys are different. Nevertheless, the 5 similar trajectories found indicate that even a naive approach such as LCS can obtain similar journeys (for route prediction purposes).

Further work in this area would involve finding a better similarity measure in combination with a trajectory clustering approach. Such approaches are, when used appropriately, able to provide predictions on the destination and route that a person is intending to take \cite{chen-2011-personal}.

\bibliography{cust-seg-traj-wifi-data} 
\bibliographystyle{ieeetr}

\end{document}